\documentclass{article}
\usepackage{arxiv}

\usepackage{cite}
\usepackage{amsmath,amssymb,amsfonts}
\usepackage{algorithmic}
\usepackage
{graphicx}
\usepackage{textcomp}
\usepackage{xcolor}
\def\BibTeX{{\rm B\kern-.05em{\sc i\kern-.025em b}\kern-.08em
    T\kern-.1667em\lower.7ex\hbox{E}\kern-.125emX}}

\usepackage[english]{babel}
\usepackage{bm}
\usepackage{url}

\usepackage{amsmath}
\usepackage{graphicx,dsfont}
\usepackage[colorlinks=true, allcolors=blue]{hyperref}

\usepackage{graphicx}
\usepackage[utf8]{inputenc}
\usepackage{graphicx}
\usepackage{amsmath, amsxtra, amsfonts, amssymb, amstext, mathtools}
\usepackage{booktabs}
\usepackage{makecell}
\usepackage{multicol}
\usepackage{multirow}
\usepackage{longtable}
\usepackage{array,xspace}

\newcommand{\gsemo}{GSEMO\xspace}

\newcommand{\ignore}[1]{}

\usepackage{color}

\begin{document}

\title{Benchmarking Algorithms for Submodular Optimization Problems Using IOHProfiler}

\author{Frank Neumann\\
Optimisation and Logistics\\
School of Computer and Mathematical Sciences\\
The University of Adelaide\\
Adelaide, Australia
\And
Aneta Neumann\\
Optimisation and Logistics\\
School of Computer and Mathematical Sciences\\
The University of Adelaide\\
Adelaide, Australia
\And
Chao Qian\\
State Key Laboratory for Novel Software Technology\\
Nanjing University\\
Nanjing, China
\And
Viet Anh Do\\
Optimisation and Logistics\\
School of Computer and Mathematical Sciences\\
The University of Adelaide\\
Adelaide, Australia
\And
Jacob de Nobel\\
Leiden Institute of Advanced Computer Science\\
Leiden University\\
Leiden, The Netherlands
\And
Diederick Vermetten\\
Leiden Institute of Advanced Computer Science\\
Leiden University\\
Leiden, The Netherlands
\And
Saba Sadeghi Ahouei\\
Optimisation and Logistics\\
School of Computer and Mathematical Sciences\\
The University of Adelaide\\
Adelaide, Australia
\And
Furong Ye\\
Leiden Institute of Advanced Computer Science\\
Leiden University\\
Leiden, The Netherlands
\And
Hao Wang\\
Leiden Institute of Advanced Computer Science\\
applied Quantum algorithms (aQa)\\
Leiden University\\
Leiden, The Netherlands
\And
Thomas B{\"a}ck\\
Leiden Institute of Advanced Computer Science\\
Leiden University\\
Leiden, The Netherlands
}
\maketitle
\thispagestyle{plain}
\pagestyle{plain}
\begin{abstract}
Submodular functions play a key role in the area of optimization as they allow to model many real-world problems that face diminishing returns. Evolutionary algorithms have been shown to obtain strong theoretical performance guarantees for a wide class of submodular problems under various types of constraints while clearly outperforming standard greedy approximation algorithms. This paper introduces a setup for benchmarking algorithms for submodular optimization problems with the aim to provide researchers with a framework to enhance and compare the performance of new algorithms for submodular problems. The focus is on the development of iterative search algorithms such as evolutionary algorithms with the implementation provided and integrated into IOHprofiler which allows for tracking and comparing the progress and performance of iterative search algorithms. We present a range of submodular optimization problems that have been integrated into IOHprofiler and show how the setup can be used for analyzing and comparing iterative search algorithms in various settings.
\end{abstract}

\section{Introduction}

Many real-world optimization problems face diminishing returns and can be modeled in terms of a submodular function~\cite{DBLP:books/cu/p/0001G14,Nemhauser:1978,vondrak2010submodularity}. Over the last 10 years, it has been shown that evolutionary algorithms (EAs) using multi-objective formulations provably obtain best possible worst-case performance guarantees for a wide range of submodular optimization problems~\cite{DBLP:journals/ec/FriedrichN15,DBLP:conf/nips/QianYZ15}. Furthermore, it has been shown that they usually also achieve better experimental performance than approximation algorithms based on greedy approaches for these problems~\cite{DBLP:conf/aaai/000100QR19,DBLP:conf/aaai/Do021,DBLP:conf/ppsn/DoN20,DBLP:journals/corr/abs-1911-11451,DBLP:conf/ppsn/NeumannN20}.

While these results show the usefulness of evolutionary algorithms for submodular problems, only a very limited number of algorithms has been explored. The focus of theoretical studies of the runtime analysis of evolutionary algorithms for submodular functions is on an evolutionary multi-objective algorithm known as \gsemo~\cite{DBLP:journals/tec/LaumannsTZ04,Giel2003} which is used in the context of Pareto optimization, i.e., the objective function and constraint function are used in a bi-objective model where the best feasible solution is returned at the end of the optimization process.
With this paper, we describe the set up of benchmarks for a competition of iterative search algorithms for submodular problems. We classify the problems into different problem categories and provide prominent combinatorial optimization problems for each category. All problems are integrated into IOHprofiler which is a tool for comparing and analyzing the performance of iterative search algorithms.

The IOHprofiler tool is a modular framework for benchmarking iterative optimization heuristics. IOHexperimenter~\cite{DBLP:journals/corr/abs-2111-04077} supports the benchmarking pipeline by providing a common interface for benchmark problems, to which a wide variety of logging functionality can be attached.
The logging enables the performance data to be recorded using common data formats, which can be loaded into the IOHanalyzer~\cite{DBLP:journals/telo/WangVYDB22} tool for interactive visualization and analysis. Since this can all be done via a GUI on a website, no additional effort is required to compare results with the provided baselines. 

We define different submodular problems that are also included in the competition on \emph{Evolutionary Submodular Optimisation} at the Genetic and Evolutionary Computation Conference (GECCO) 2022 and 2023.
Let $c(x)$ be the cost of a solution $x$ and $f(x)$ the value of the (submodular) function applied to $x$. 
Our cost functions are either linear as in the classical knapsack problem or are given based on tail bounds used to evaluate chance constraints that are imposed on submodular functions when dealing with stochastic settings~\cite{DBLP:conf/gecco/XieHAN019,DBLP:conf/ppsn/NeumannN20,DBLP:conf/ppsn/NeumannXN22}.
All settings consider the maximization of a given submodular function $f$ under a (possible) cost constraint $c(x) \leq B$, where $B$ is a given constraint bound. 

Our setup divides the class of submodular problems into monotone and non-monotone submodular problems. For monotone submodular optimization problems, the function value does not decrease when adding extra elements. Here the constraint imposes the difficulty as all solutions have to meet the cost required, i.e., $c(x) \leq B$ is required for any feasible solution $x$. The example problems that we integrate are the maximum coverage problem in graphs~\cite{khuller1999budgeted} and the maximum influence problem in social networks~\cite{DBLP:conf/kdd/KempeKT03}. 

For non-monotone submodular optimization problems, it is not necessary to impose a cost constraint as such problems are already hard without the constraint. Examples include the classical maximum cut problem in graphs~\cite{DBLP:reference/algo/Newman08} and the packing while traveling problem~\cite{DBLP:journals/eor/PolyakovskiyN17} which constitutes the packing component of the multi-component traveling thief problem~\cite{DBLP:conf/cec/BonyadiMB13}. The maximum cut problem does not come with an additional constraint as each selection of nodes defines one partition whereas the second portion is given by the remaining nodes and the edges crossing the partitions constitute the edges of the cut. For the packing while traveling problem which is a submodular and non-linear knapsack problem, we impose a standard knapsack constraint that limits the sum of the weights of the selected items.

The paper is structured as follows. In Section~\ref{sec2}, we describe the setup of submodular optimization and give an overview of the IOHprofiler framework. We describe monotone submodular benchmark problems in Section~\ref{sec3}, and non-monotone submodular benchmark problems in Section~\ref{sec4}. Section~\ref{sec5} gives our integration of these problems into IOHprofiler and showcases some results for comparing algorithms on the integrated problems. Finally, we finish with some concluding remarks.

\section{Preliminaries}
We now describe the setup on submodular optimization and provide an introduction into IOHprofiler.
\label{sec2}

\subsection{Submodular Optimization}
Many real-world problems can be formulated in terms of optimizing a submodular function under a given set of constraints. We consider discrete optimization problems where the goal is to select a feasible subset of a given set of elements.

Let $V = \{v_1, \ldots, v_n\}$ be a ground set of $n$ elements.
 A function $f \colon 2^V \rightarrow \mathds{R}_{\geq 0}$ is called submodular iff for every $S, T \subseteq V$ with $S \subseteq T$ and $v \not \in T$
$$f(S \cup \{v\}) - f(S) \geq f(T \cup \{v\}) - f(T)$$
holds.
A function $f$ is called monotone if $f(A) \leq f(B)$ for any $A \subseteq B \subseteq V$.
Throughout this paper, we encode solutions as binary strings of length $n$ and identify a set $X = \{v_i \mid x_i=1\}$ by the elements $v_i$ where $v_i \in X \Leftrightarrow x_i = 1$.
The set of feasible solutions is defined in terms of a cost constraint in this paper. 
We denote by $c(x)$ the cost of a solution $x$ and a solution $x$ is feasible iff $c(x) \leq B$ holds. 
We denote by $F \subseteq 2^V$ the set of feasible solutions, and define the feasible search points/solutions as $F = \{x \in \{0,1\}^n \mid c(x) \leq B\}$.

We consider problems where the goal is to find a feasible solution $x^*$ with 

$$
x^* = \arg \max_{x \in F} f(x).
$$

All problems considered in this paper are submodular and we distinguish between monotone and non-monotone problems.
The function $f$ is dependent on the problem at hand and we will describe important optimization problems that fit the submodular formulation in Section~\ref{sec3} and \ref{sec4}.

\subsection{Types of constraints}
Monotone submodular functions without any constraint are trivial to optimize as selecting all elements from the given set yields an optimal solution. Problems are usually defined in terms of an objective function and a given set of constraints that limit the resources available to achieve high-quality solutions.

We consider different types of deterministic and stochastic constraints.
As deterministic constraints, we consider linear constraints where we have $c(x) = \sum_{i=1}^n c(v_i) x_i \leq B$ and $c(v_i) \geq 0$. This includes the special case of a uniform constraint, i.e., where $c(v_i) = 1, 1 \leq i \leq n$, holds.

We also consider surrogate constraint functions used in the area of chance constraints~\cite{DBLP:journals/corr/abs-1911-11451,DBLP:conf/ppsn/NeumannN20}. Here the cost $c(x)$ of a solution $x$ is stochastic and the constraint is given as $Pr(c(x)>B) \leq \alpha$, where $\alpha \in ]0, 1/2]$ is an upper bound on the probability that the constraint is violated. Evaluating whether a given solution fulfills a given chance constraint is computationally expensive in general, and surrogate approaches based on tail inequalities such as Chebyshev's inequality and Chernoff bounds provide a suitable alternative.

We consider the case where $c(v)$ is chosen uniformly at random in $[a(v) - \delta, a(v) + \delta]$ where $a(v)$ is the expected cost of an element $v$ and $\delta$ is a parameter determining the uncertainty. Let $a({x}) = \sum_{i=1}^n a(v_i) x_i$ be the expected cost of solution ${x}$ and $v({x}) = |{x}|_1 \cdot \delta^2/3$ be its variance. Based on tail bounds used to evaluate chance constraints~\cite{DBLP:conf/aaai/DoerrD0NS20,DBLP:conf/ppsn/NeumannN20}, we consider the following cost functions to make sure that the chance constraint is met.

We use the cost function
\begin{eqnarray*}
c_{Cheby}({x}) & = & a({x}) + \sqrt{\frac{1-\alpha}{\alpha}\cdot v({x})}\\
& = & a({x}) + \delta \cdot \sqrt{\frac{1-\alpha}{3\alpha} \cdot |{x}|_1}
\end{eqnarray*}
based on Chebyshev's inequality~\cite{DBLP:books/cu/MotwaniR95}. The chance constraint is met if $c_{Cheby}({x}) \leq B$ holds.

We use the cost function

$$c_{Cher}({x}) = a({x}) + \delta \cdot \sqrt{\ln(1/\alpha) \cdot 2|{x}|_1}$$
based on Chernoff bounds~\cite{DBLP:books/cu/MotwaniR95}.
The chance constraint is met if $c_{Cher}({x}) \leq B$ holds. 
We will investigate the mentioned type of constraints together with the submodular optimization problems outlined in Section~\ref{sec3} and \ref{sec4}.

\subsection{IOHProfiler}
IOHprofiler\footnote{\url{https://iohprofiler.github.io/}} is a framework for benchmarking iterative optimization heuristics. It consists of two main components: IOHexperimenter, which provides an interface between benchmark problems, algorithms, and logging; and IOHanalyzer, which enables visualization and analysis of the recorded performance data. 

IOHexperimenter\footnote{\url{https://github.com/IOHprofiler/IOHexperimenter}} provides access to a wide variety of benchmark suites (currently only single-objective, noiseless problems are included). These problems can contain arbitrary constraints, which enables us to integrate the submodular problems directly. While the implementation is done in C++, a Python interface is available, which contains the complete set of features available in C++. 

The data resulting from running an algorithm using IOH\-experimenter can be directly used with IOHanalyzer, which can be accessed via a web-based GUI\footnote{\url{iohanalyzer.liacs.nl}. Also available as an R-package directly from CRAN.}. With IOHanalyzer, performance trajectories can be visualized from both a fixed-budget and fixed-target perspective, as well as on a higher level of aggregation, e.g.,~through empirical density functions. On IOHanalyzer, the data from the baseline algorithms has been made available for comparison. This data can be accessed by selecting `IOH' as the repository, and then `Submodular\_SUITENAME' as the data source. 

\section{Monotone Submodular Benchmark Problems}
\label{sec3}
Monotone submodular optimization problems contain some prominent example problems and have widely been studied in the literature. For various types of constraints, it has been shown that greedy algorithms and evolutionary multi-objective algorithms achieve the best possible worst-case performance guarantees.
\subsection{Maximum Coverage}
The maximum coverage problem is a classical optimization problem on graphs~\cite{khuller1999budgeted}.
Given an undirected weighted graph $G=(V,E, c)$ with costs $c \colon V \rightarrow \mathds{R}_{\geq 0}$ on the vertices.
We denote by $N(V')= \{v_i \mid \exists e \in E: e\cap V' \not = \emptyset \wedge e \cap v_i \not = \emptyset\}$ the set of all nodes of $V'$ and their neighbors in $G$.

For a given search point ${x} \in \{0,1\}^n$ where $n = |V|$, we have $V'({x}) = \{v_i \mid x_i=1\}$ and $c({x}) = \sum_{v \in V'({x})} c(v)$.

\subsubsection{Deterministic Setting}
In the deterministic setting, the goal is to maximize
$$f({x}) = |N(V'({x}))|$$
under the constraint that $c({x}) \leq B$ holds. 

The fitness of a search point ${x}$ is given as the $2$-dimensional vector
$g({x}) = (f'({x}), c({x}))$ where
$$f'({x}) = \begin{cases}
f({x}) & c({x}) \leq B\\
B-c({x}) & c({x}) >B\\
\end{cases}
$$
This implies that each infeasible solution has a negative fitness value whereas each feasible solution has a non-negative one.
\paragraph{Experimental setting}
We integrated different graphs into IOHprofiler that can be used for experimentation. Note that the framework is flexible in the sense that users can add additional graphs if required. 
Example graphs include frb-graphs\footnote{\url{https://github.com/dynaroars/npbench/tree/master/instances/vertex\_cover/benchmarks}} with up to $760$ nodes which have frequently been used for covering problems.
For costs and budgets for benchmarking, we use
\begin{itemize}
    \item uniform: $c(v)=1$, $\forall v\in V$, $B = 10$
    \item linear-degree: $c(v)=1+deg(v)$, $B=500$
    \item quadratic-degree: $c(v) = (1+deg(v))^2$, $B = 40000$
\end{itemize}
where $deg(v)$ denotes the degree of node $v$.

\subsubsection{Chance constrained setting}

We take the cost of a node for the given benchmark instance as expected cost and consider the uncertainty parameterized by $\delta$.

\paragraph{Cost function based on Chebyshev' inequality}
The fitness of a search point ${x}$ using $c_{Cheby}$ is given as the 2-dimensional vector
$g_{Cheby}({x}) = (f_{Cheby}'({x}), c_{Cheby}({x}))$ where
$$f_{Cheby}'({x}) = \begin{cases}
f({x}) & c_{Cheby}({x}) \leq B\\
B-c_{Cheby}({x}) & c_{Cheby}({x}) >B\\
\end{cases}
$$

\paragraph{Cost function based on Chernoff bounds}
The fitness of a search point ${x}$ using $c_{Cher}$ is given as the 2-dimensional vector
$g_{Cher}({x}) = (f_{Cher}'({x}), c_{Cher}({x}))$ where
$$f_{Cher}'({x}) = \begin{cases}
f({x}) & c_{Cher}({x}) \leq B\\
B-c_{Cher}({x}) & c_{Cher}({x}) >B\\
\end{cases}
$$

\paragraph{Experimental setting}
We use the graphs and bounds as for the deterministic case. Special chance constraint parameters are as follows.
\begin{itemize}
\item uniform: We use $a(v) = 1$, $\forall v \in V$, and $\delta \in \{0.5, 1\}$, $\alpha \in \{0.1, 0.01, 0.001\}$. 
\item linear-degree: We use $a(v) = 1 + deg(v)$, and $\delta \in \{20, 40\}$, $\alpha \in \{0.1, 0.01, 0.001\}$.
\end{itemize}

\subsection{Maximum Influence}
The maximum influence problem in social networks is an important submodular optimization problem that has been widely studied in the literature from various perspectives \cite{DBLP:conf/kdd/KempeKT03,DBLP:conf/wsdm/GoyalBL10, DBLP:journals/datamine/WangCW12}.
Let a directed graph $G(V, E)$ represent a social network, where each node is a user and each edge $(u, v)\in E$ has a probability $p_{u,v}$ representing the strength of influence from user $u$ to $v$.  

A fundamental propagation model is independence cascade. Starting from a seed set $X$, it uses a set $A_t$ to record the nodes activated at time $t$, and at time $t + 1$, each inactive neighbor $v$ of $u\in A_t$ becomes active with probability $p_{u,v}$. This process is repeated until no nodes get activated at some time. The set of nodes activated by propagating from $X$ is denoted as $IC(X)$, which is a random variable. 

The goal is to maximize the expected value of $IC(X)$.
Note that the computation of the expected value is done by running a simulation of the influence process several times and averaging its results. In this sense, the computation of the objective function value is stochastic.

For a given search point ${x} \in \{0,1\}^n$ where $n = |V|$, we have $V'({x}) = \{v_i \mid x_i=1\}$ and $c({x}) = \sum_{v \in V'({x})} c(v)$.

\subsubsection{Deterministic Setting}

In the deterministic setting, the goal is to maximize the expected number of nodes activated by propagating from $V'({x})$, i.e.,
$$f({x}) =\mathbb{E}[|IC(V'({x}))|]$$
under the constraint that $c({x}) \leq B$ holds. 

The fitness of a search point ${x}$ is given as the 2-dimensional vector
$g({x}) = (f'({x}), c({x}))$ where
$$f'({x}) = \begin{cases}
f({x}) & c({x}) \leq B\\
B-c({x}) & c({x}) >B\\
\end{cases}
$$

\paragraph{Experimental setting} Real-world data sets have been downloaded from SNAP networks\footnote{\url{http://snap.stanford.edu/data/index.html}}. After preprocessing, we get a directed graph with several nodes and edges. We determine the probability of one edge from $v_i$ to $v_j$ by $\frac{weight(v_i,v_j )}{indegree(v_j)}$. For estimating the influence spread, i.e., the expected number of active nodes, we simulate the diffusion process multiple times independently and use the average as an estimation. 

As an example, we consider the data set ego-Facebook. For costs and budgets, we use the following settings:
\begin{itemize}
    \item uniform: $c(v)=1$, $\forall v\in V$, $B = 10, 20, 50, 100$.
    \item linear-degree: $c(v)=1+outdegree(v)$, $\forall v\in V$, $B = 200, 400, 1000, 2000.$
\end{itemize}

\subsubsection{Chance constrained setting}
We have $f({x})=\mathbb{E}[|IC(V'({x}))|]$ as defined in the deterministic setting. 
We take the cost of a node for the given benchmark instance as expected cost and consider the uncertainty parameterized by $\delta$.

\paragraph{Cost function based on Chebyshev' inequality}
As done for the maximum coverage problem, we use the cost function $c_{Cheby}$
based on Chebyshev's inequality.

The fitness of a search point ${x}$ using $c_{Cheby}$ is given as the 2-dimensional vector
$g_{Cheby}({x}) = (f_{Cheby}'({x}), c_{Cheby}({x}))$ where
$$f_{Cheby}'({x}) = \begin{cases}
f({x}) & c_{Cheby}({x}) \leq B\\
B-c_{Cheby}({x}) & c_{Cheby}({x}) >B\\
\end{cases}
$$

\paragraph{Cost function based on Chernoff bounds}
As done for the maximum coverage problem, we use the cost function $c_{Cher}$
based on Chernoff bounds.
The chance constraint is met if $c_{Cher}({x}) \leq B$ holds. 

The fitness of a search point ${x}$ using $c_{Cher}$ is given as the 2-dimensional vector
$g_{Cher}({x}) = (f_{Cher}'({x}), c_{Cher}({x}))$ where
$$f_{Cher}'({x}) = \begin{cases}
f({x}) & c_{Cher}({x}) \leq B\\
B-c_{Cher}({x}) & c_{Cher}({x}) >B\\
\end{cases}
$$

\paragraph{Experimental setting}

Graphs and bounds as for the deterministic case are available in IOHprofiler. The special chance constraint parameters are as follows.
\begin{itemize}
\item uniform: Use $a(v) = 1$, $\forall v \in V$, and $\delta \in \{0.5, 1\}$, $\alpha \in \{0.1, 0.01, 0.001\}$. 
\item linear-degree: Use $a(v) = 1 + outdegree(v)$, and $\delta \in \{20, 40\}$, $\alpha \in \{0.1, 0.01, 0.001\}$.
\end{itemize}

\section{Non-monotone Submodular Benchmark Problems}
\label{sec4}
So far, we considered monotone submodular optimization problems where the difficulty occurs through the given constraint. Optimizing a non-monotone submodular function without any additional constraint is already NP-hard as for example the well-known maximum cut problem in graphs can be stated in terms of optimizing a non-monotone submodular function. The maximum cut problem is also our first benchmark problem for the category of a non-monotone submodular function, and we consider this problem in its classical way without any additional constraint.
\subsection{Maximum Cut}
The maximum cut problem~\cite{DBLP:reference/algo/Newman08} is a classical NP-hard problem and can be defined as follows.
Given an undirected weighted graph $G=(V,E, w)$ with weights $w \colon E \rightarrow \mathds{R}_{\geq 0}$ on the edges, the goal is to select a set $V_1 \subseteq V$ such that the sum of the weight of edges between $V_1$ and $V_2 = V \setminus V_1$ is maximal.

For a given search point ${x} \in \{0,1\}^n$ where $n = |V|$, we have $V_1({x}) = \{v_i \mid x_i=1\}$ and  $V_2({x}) = \{v_i \mid x_i=0\}$. Let
$C({x}) = \{e \in E \mid e \cap V_1({x}) \not = \emptyset \wedge e \cap V_2({x}) \not = \emptyset\}$ be the cut of a given search point ${x}$.
The goal is to maximize
$$f'({x}) = \sum_{e \in C({x})} w(e).$$
Note that every search point in $\{0,1\}^n$ is feasible and there is therefore no penalty or second objective for treating potentially infeasible solutions.

\paragraph{Experimental setting}
Different G-Set graphs\footnote{https://web.stanford.edu/~yyye/yyye/Gset/} have been incorporated into IOHprofiler to carry out experimental investigations.

\subsection{Packing While Traveling}

The packing while traveling (PWT) problem~\cite{DBLP:journals/eor/PolyakovskiyN17} is a non-monotone submodular optimization problem which is obtained from the traveling thief problem (TTP)~\cite{DBLP:conf/cec/BonyadiMB13} when the route is fixed.  

The input is given as $n+1$ cities with distances $d_i$, $1\leq i \leq n$, from city $i$ to city $i+1$. 
Each city $i$, $1 \leq i \leq n$, contains a set of items $M_i \subseteq M$, $|M_i| = m_i$. Each item $e_{ij} \in M_i$, $1 \leq j \leq m_i$, has a positive integer profit $p_{ij}$ and weight $w_{ij}$.
A fixed route $N = (1, 2, ..., n+1)$ is traveled by a vehicle with velocity $v \in [v_{min},v_{max}]$. We denote by $x_{ij} \in \{0, 1\}$ the variable indicating whether or not item $e_{ij}$ is chosen in a solution 
$$x = (x_{11},x_{12},...,x_{1m_1},x_{21},...,x_{nm_n}) \in \{0,1\}^m,$$ 
where $m=\sum^{n}_{i=1}m_i$. The total benefit of selecting a subset of items selected by ${x}$ is given as
\begin{equation*}\label{bpt}
	PWT({x}) = P({x}) - R \cdot T({x}),
\end{equation*}
where $P({x})$ is the total profit of the selected items and $T({x})$ is the total travel time for the vehicle carrying the selected items. Formally, we have
\begin{equation*}\label{profit}
	P({x}) = \sum\limits_{i=1}^n \sum\limits_{j=1}^{m_i} p_{ij}x_{ij}
\end{equation*}
and
\begin{equation*}\label{cost}
	T({x}) = \sum\limits_{i=1}^n \frac{d_i}{v_{max} - \nu\sum\limits_{k=1}^i\sum\limits_{j=1}^{m_k} w_{kj}x_{kj}}
\end{equation*}

Here, $\nu = \frac{v_{max}-v_{min}}{B}$ is a constant defined by the input parameters, where $B$ is the capacity of the vehicle. The problem is already NP-hard without any additional constraint~\cite{DBLP:journals/eor/PolyakovskiyN17}, but often considered with a typical knapsack constraint given as
\begin{equation*}\label{weight}
	c({x}) = \sum\limits_{i=1}^n \sum\limits_{j=1}^{m_i} w_{ij}x_{ij} \leq B.
\end{equation*}

As fitness functions, we use
 $g({x}) = (f'({x}), c({x}))$ with
$$f'({x}) = \begin{cases}
PWT({x}) & c({x}) \leq B\\
B-c({x})-R \cdot T(v_{\min}) & c({x}) >B\\
\end{cases}
$$
where $T(v_{\min})=\frac{1}{v_{\min}}\cdot \sum\limits_{i=1}^n d_i$ is the travel time at speed $v_{\min}$. 

\paragraph{Experimental setting}
Instances of the TTP\footnote{\url{https://cs.adelaide.edu.au/~optlog/TTP2017Comp/}} have been incorporated into IOHprofiler and the permutation to be used for PWT is fixed as $N=id=(1, \ldots, n)$. The start for packing while traveling is at city $1$ and ends at city $n+1:=1$.

\section{Submodular Problems in IOHProfiler}
\label{sec5}
The submodular problems presented in this paper have been implemented in IOHexperimenter. IOHexperimenter defines constrained optimization problems as $F_c = F \circ C$, where $C$ denotes a set of $k$ constraints of the general form $C_i: \mathbb{R}^n \to \mathbb{R}$. The submodular problems are implemented with the constraints $C_i: $\{0,1\}$^n \to \mathbb{R}_{\geq 0}$, which penalize the objective function value as: 
\begin{equation*}
f_c(x) =
\begin{cases}
    \sum_{i=1}^k w_i(C_i(x))^{\alpha_i}, & \sum_{i=1}^k C_i(x) > 0\\
    f(x) & \text{otherwise}
\end{cases}
\end{equation*}
$C_i(x) > 0$ when there is constraint violation, and $w_i$ and $\alpha_i$ are predefined weight and exponent parameters. For each of the submodular problems, only a single constraint function is implemented (i.e., $k=1$), which is defined according to the cost functions described in the previous Sections~\ref{sec3}-\ref{sec4}. The implementation is available on Github\footnote{\url{https://github.com/IOHprofiler/IOHexperimenter/tree/master/include/ioh/problem/submodular}}, and the problems have been integrated as one of the available suites in IOHexperimenter.

\subsection{Algorithms}
To provide a future baseline for the introduced submodular problem suite, we compare the following $12$ algorithms. Most of the tested algorithms, except $(1+(\lambda,\lambda))$~EA$_{>0}$, random search, and univariate marginal distribution algorithm,  are mutation-only methods. The applied mutation flips $\ell$ distinct bits selected uniformly at random. Note that we force $\ell >0$ while sampling $\ell$ in practice, following the suggestion in~\cite{DBLP:journals/corr/abs-1812-00493}. 

We briefly introduce the algorithms as below and provide our implementation in GitHub\footnote{\url{https://github.com/IOHprofiler/IOHalgorithm}}. 

\begin{itemize}
    \item \textbf{$(1+1)$~EA$_{>0}$}: The $(1+1)$~EA$_{>0}$ using the standard bit mutation with a static mutation rate $p=1/n$. The standard bit mutation samples $\ell$, the number of distinct bits to be flipped, from a conditional binomial distribution Bin$_{>0}(n,p)$.
    \item \textbf{$(1+1)$~fast genetic algorithm (fast GA)}: The $(1+1)$ fast GA differs from the $(1+1)$~EA by sampling $\ell$ from a power-law distribution with $\beta=1.5$~\cite{DBLP:conf/gecco/DoerrLMN17}. The power-law distribution is a heavy-tailed distribution, and its probability of sampling large $\ell > 1$ is higher, compared to the standard bit mutation with $p=1/n$.
    \item \textbf{$(1+(\lambda,\lambda))$~EA$_{>0}$}: The $(1+\lambda)$~EA$_{>0}$ with self-adaptive $\lambda$ proposed in~\cite{DBLP:journals/tcs/DoerrDE15}. The algorithm applies the standard bit mutation and a biased or parameterized uniform crossover, where the mutation rate and the crossover probability depend on the value of $\lambda$. We implement in this paper the $(1+(10,10))$~EA$_{>0}$ with an initial $\lambda = 10$.
    \item \textbf{$(1+10)$~2rate-EA$_{>0}$}: The $(1+10)$~EA$_{>0}$ using standard bit mutation with self-adaptive mutation rates. The self-adaptive technique is proposed in~\cite{DBLP:conf/gecco/DoerrGWY17}.
    \item \textbf{$(1+10)$~normEA$_{>0}$}: The $(1+10)$~EA$_{>0}$ using the normalized bit mutation~\cite{DBLP:conf/cec/YeDB19} with self-adaptive $r$, which samples $\ell$ from a normal distribution N$(r,r(1-r/n))$.
    \item \textbf{$(1+10)$~varEA$_{>0}$}: The algorithm also applied the normalized bit mutation. However, it controls the variance of the normal distribution using a factor $F^c$, where $c$ is the evaluation time since the best-found fitness has been updated, and we set $F=0.98$. Also from~\cite{DBLP:conf/cec/YeDB19}.
    \item \textbf{greedy hill climber (gHC)}: The $(1+1)$ gHC flips one bit, going through the bit-string from left to right, in each iteration. It updates the parent when the offspring obtains fitness at least as good as its parent. 
    \item \textbf{random search}: The random search samples new solutions uniformly at random iteratively.
    \item \textbf{randomized local search (RLS)}: The RLS differs from the $(1+1)$~EA$_{>0}$ by flipping exactly $\ell=1$ bit in each iteration.
    \item \textbf{simulated annealing (sa-auto)}: The SA algorithm with automatic settings based on the problem dimension. For the start temperature, the probability of accepting a solution that is $1$ worse than the current solution is $0.1$, and this value is $1/\sqrt{n}$ for the end temperature. Discussions about the SA can be found in~\cite{nourani1998comparison}.
    \item \textbf{simulated annealing with iterative restart (sars-auto)}: The algorithm applies iterative restarts for the SA. We assign different function evaluation budgets to each restarting round.
    \item \textbf{univariate marginal distribution algorithm (UMDA)}: UMDA~\cite{DBLP:journals/ec/Muhlenbein97} maintains a population of $s$ solutions and uses the best $s/2$ solutions to estimate marginal distributions for each variable. It samples new populations based on the marginal distributions and updates the distributions iteratively. We set $s=50$ in our experiments. 
\end{itemize}

\subsection{Assessment and Comparison Using IOHprofiler}

To ease the assessment of new algorithms on these submodular problems, we provide performance data from a set of 12 baseline algorithms directly on IOHanalyzer. As the postprocessing tool of the IOHprofiler, IOHanalyzer~\cite{DBLP:journals/telo/WangVYDB22} provides access to a wide range of interactive visualization and analysis methods, which can be used directly from the GUI at \url{iohanalyzer.liacs.nl}. Performance data generated using IOHexperimenter can be loaded and compared to the available baselines from either a fixed-budget or fixed-target perspective. 
We show a subset of available figures here, but recommend the reader to interactively explore the data by loading it via the IOHanalyzer GUI (data repositories \verb|Submodular_MaxCoverage| and \verb|Submodular_MaxCut|).
We present in this section the experimental results, regarding fixed-target results, i.e., \emph{Expected Running Time (ERT)} and \emph{aggregated Empirical Cumulative Distribution Function (ECDF) curve}, and fixed-budget results, i.e., the \emph{glicko2 ranking}, for the assessments of the algorithms' performance, and a heatmap of pairwise comparisons regarding the best-found function values after a set of evaluation times. The definitions of ERT and ECDF can be found in~\cite{DBLP:journals/telo/WangVYDB22}, and the glicko2 ranking system makes use of chess ranking systems to order the performance of optimization algorithms~\cite{vevcek2014chess}. We take the maximum coverage and maximum cut functions as example problems to show the results for a monotone and non-monotone submodular optimization problem. All results are based on $30$ runs per instance using 100,000 fitness evaluations per run.

\subsubsection{Results of Maximum Coverage}
\begin{figure}
    \centering
    \includegraphics[width=0.48\textwidth]{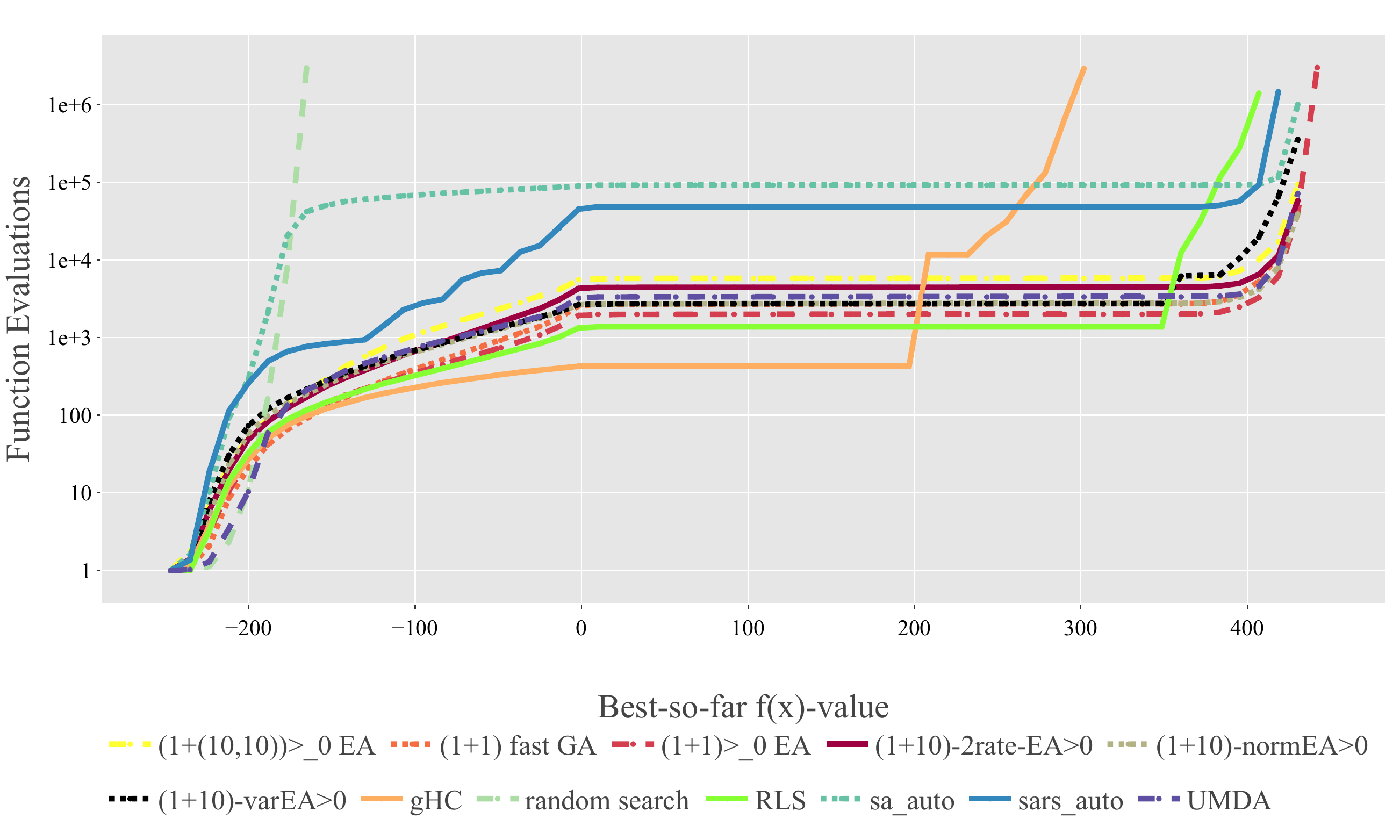}
    \caption{ERT values of 12 algorithms on a maximum coverage problem instance. Negative fitness values correspond to infeasible solutions.}
    \label{fig:ert_maxcoverage}
\end{figure}

\begin{figure}
    \centering
    \includegraphics[width=0.48\textwidth]{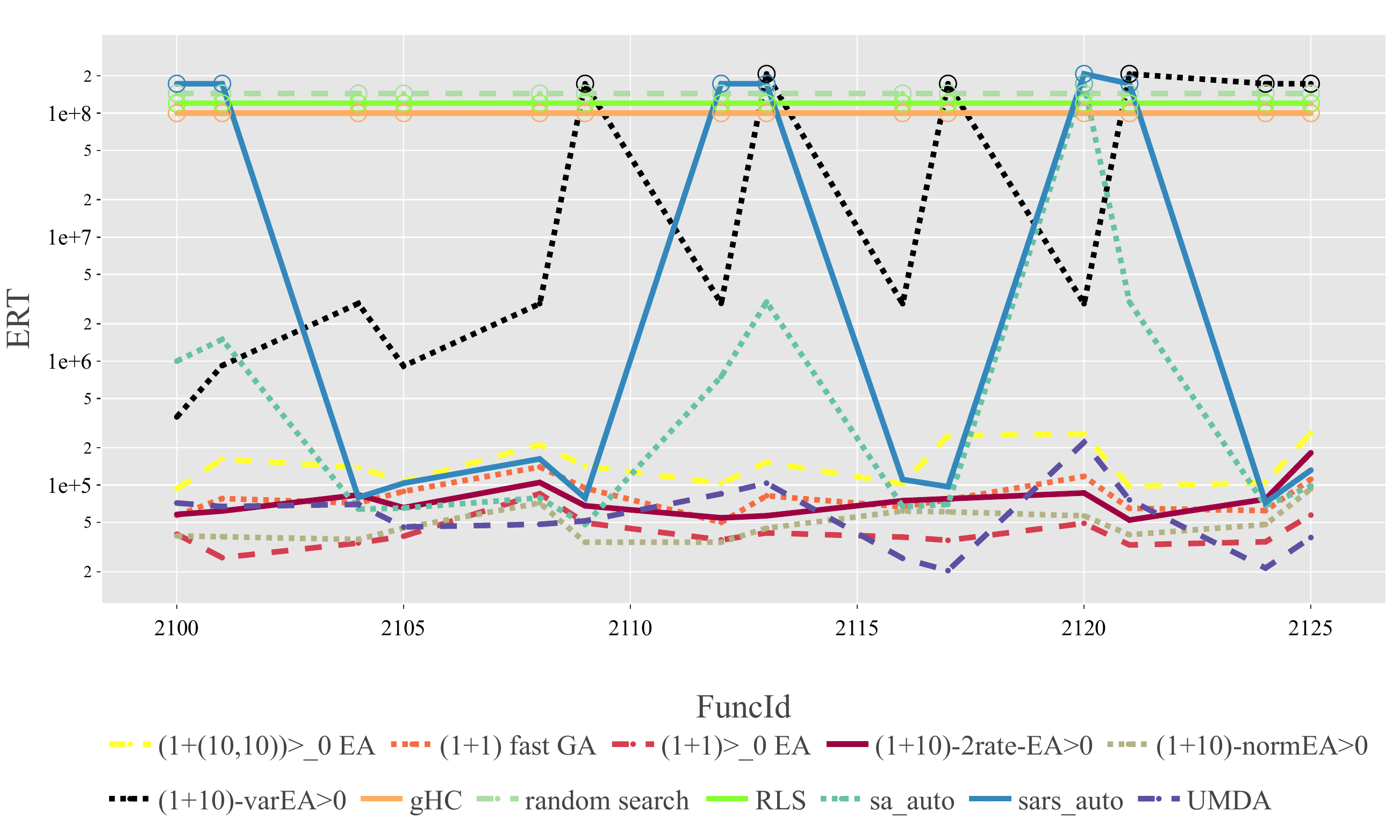}
    \caption{ERT values of 12 algorithms for the $450$-dimensional maximum coverage problem instances. The targets are selected to be the $0.02$ quantile of the final target (i.e., fitness) found by the best algorithm. Circles indicate that algorithms can not hit the corresponding target with the given budget.}
    \label{fig:ert_rank}
\end{figure}

We plot in Figure~\ref{fig:ert_maxcoverage} the ERT values of the tested algorithms for an instance of the maximum coverage function. We can see that the gHC needs relatively few function evaluations to find a decent feasible solution, but it fails to find solutions of higher quality. We also note the clear inability of random search to find any feasible solutions, while the EA variants all show similar levels of performance to UMDA. In addition, we plot in Figure~\ref{fig:ert_rank} the ERT values regarding the targets that are the 0.02 quantile of targets found by the best algorithm for multiple maximum coverage problem instances. We observe that random search, gHC, RLS, and the $(1+10)$~varEA can not find the corresponding targets with the given function evaluation budgets for all the problem instances, while the other EA variants and UMDA present relatively promising performance. The former subset of algorithms emphasizes local search by flipping an exact number of bits in mutation, while the latter ones favour global search by considering different numbers of flipping bits.  It is also interesting to notice that the performance of the two SA algorithms shows biased behavior for different problem instances.

\begin{figure}
    \centering
    \includegraphics[width=0.48\textwidth]{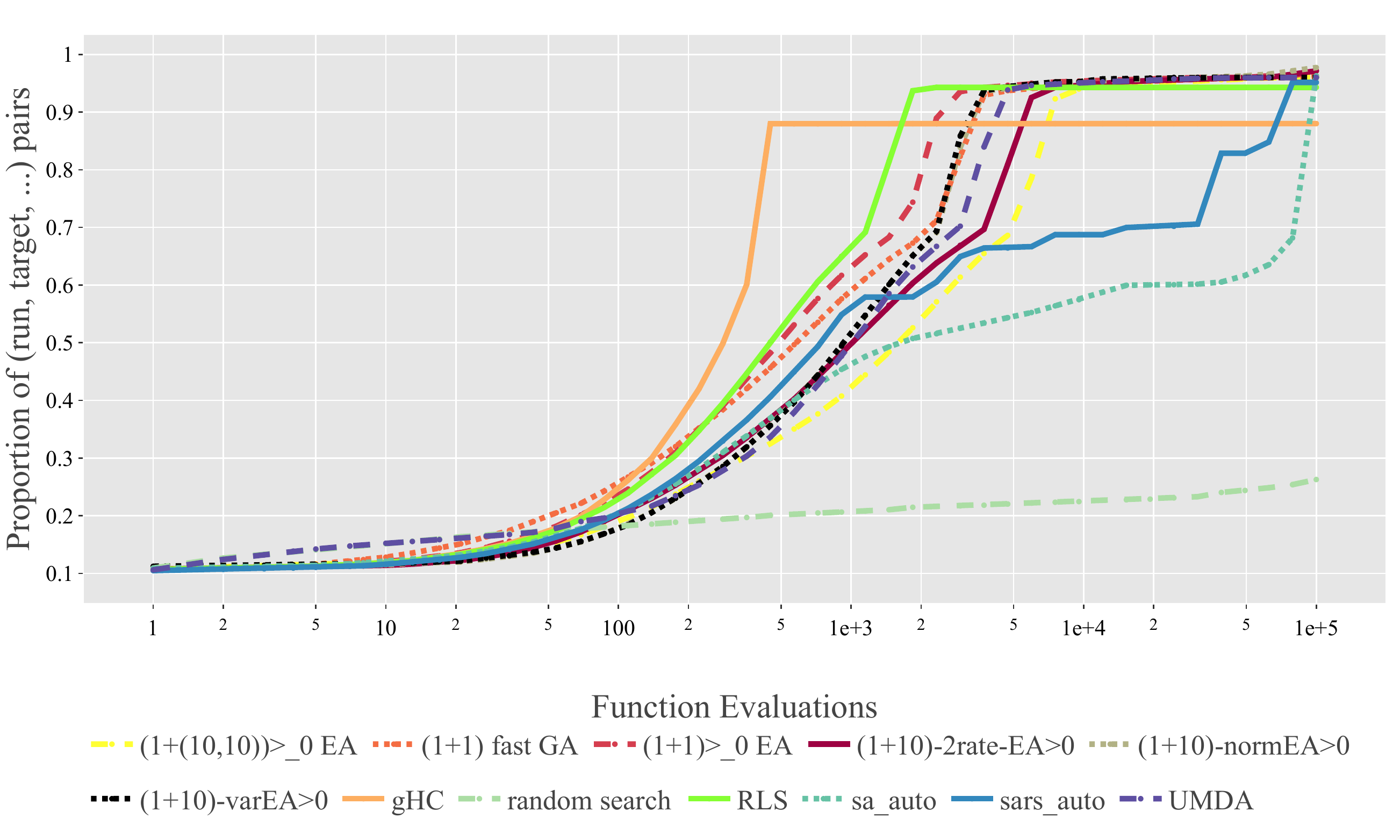}
    \caption{Empirical cumulative distribution function of 12 algorithms aggregated over all $450$-dimensional maximum coverage problem instances. Targets are $25$ linearly spaced points between the best and worst found fitness values.}
    \label{fig:ecdf}
\end{figure}

In addition to this per-function view, we can aggregate the performance of multiple functions into a single plot to get an overview of the global behavior of the selected algorithms. This can be achieved using the aggregated ECDF curve, as is shown in Figure~\ref{fig:ecdf}. This figure clearly shows the fast initial convergence of the gHC, which starkly contrasts with the slower start of the SA methods. However, gHC converges to a small ECDF value due to the incapability of obtaining the function values as good as other algorithms. Similar performance can be observed for RLS, which presents fast initial convergence but is outperformed by the EA variants within the maximal given budget. Overall, the differences between the EA variants are rather small, and the $(1+1)$~EA$_{>0}$ obtains the highest area under the curve. 

\begin{figure}
    \centering
    \includegraphics[width=0.48\textwidth]{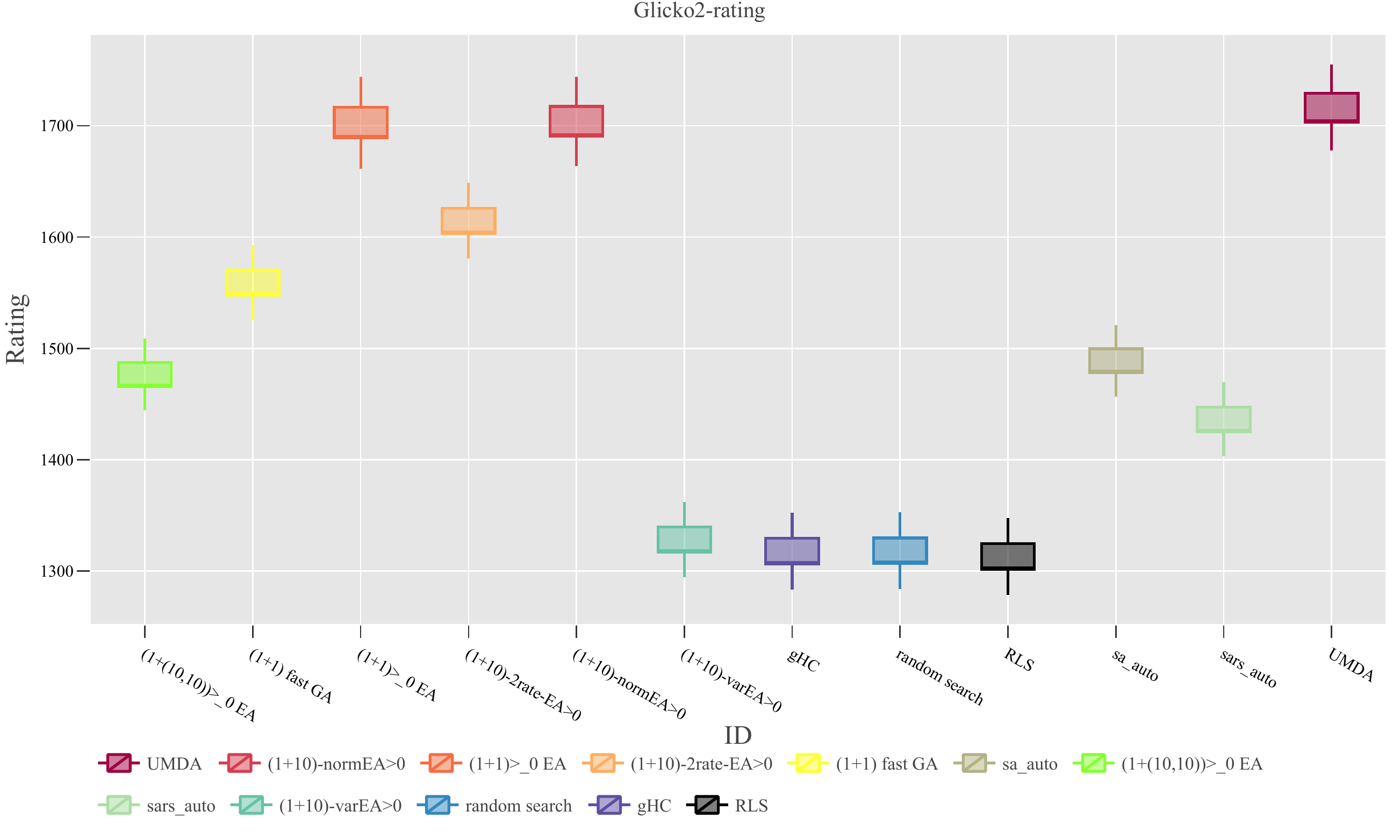}
    \caption{Glicko2 ranking of 12 algorithms over all maximum coverage problem instances, achieved by using final obtained fitness values to determine winner for $25$ pairwise games for each problem instance and each pair of algorithms. }
    \label{fig:glico2}
\end{figure}

For the fixed-budget results, we use the glicko2 ranking system to aggregate the performance over all maximum coverage problem instances. 
This is achieved by considering each function as a game where the final function value reached determines which algorithm wins the ``game''. By sampling $25$ games per function for each pair of algorithms, we get an aggregated ranking as shown in Figure~\ref{fig:glico2}. We observe that the $(1+1)$~EA$_{>0}$ obtains the highest rank, which shows identical assessment to our results of ECDF in Figure~\ref{fig:ecdf}.

\subsubsection{Results of Maximum Cut} 
\begin{figure}
    \centering
    \includegraphics[width=0.48\textwidth]{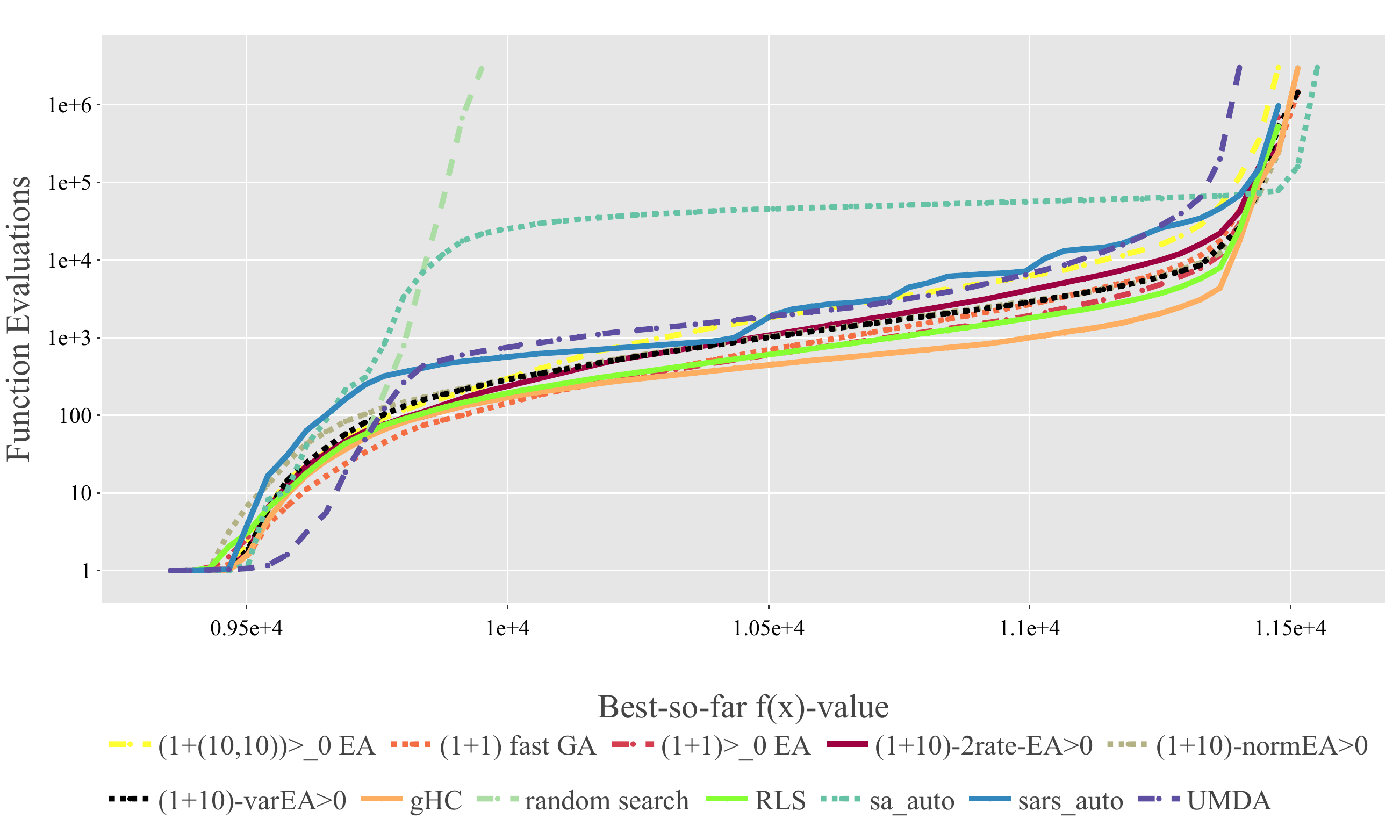}
    \caption{ERT values of 12 algorithms on a maximum cut problem instance. Negative fitness values correspond to infeasible solutions.}
    \label{fig:ert_maxcut}
\end{figure}

We plot in Figure~\ref{fig:ert_maxcut} the ERT values of the algorithms for an instance of the maximum cut function. Still, random search is incapable of searching for promising solutions, compared to the other algorithms. Meanwhile, UMDA is outperformed by the other algorithms except for random search.
However, gHC presents a fast initial convergence, and its best found fitness values are close to the ones obtained by the EA variants.
In addition, the sa-auto obtains the best fitness value for the plotted problem instance. 

\begin{figure}
    \centering
    \includegraphics[width=0.48\textwidth]{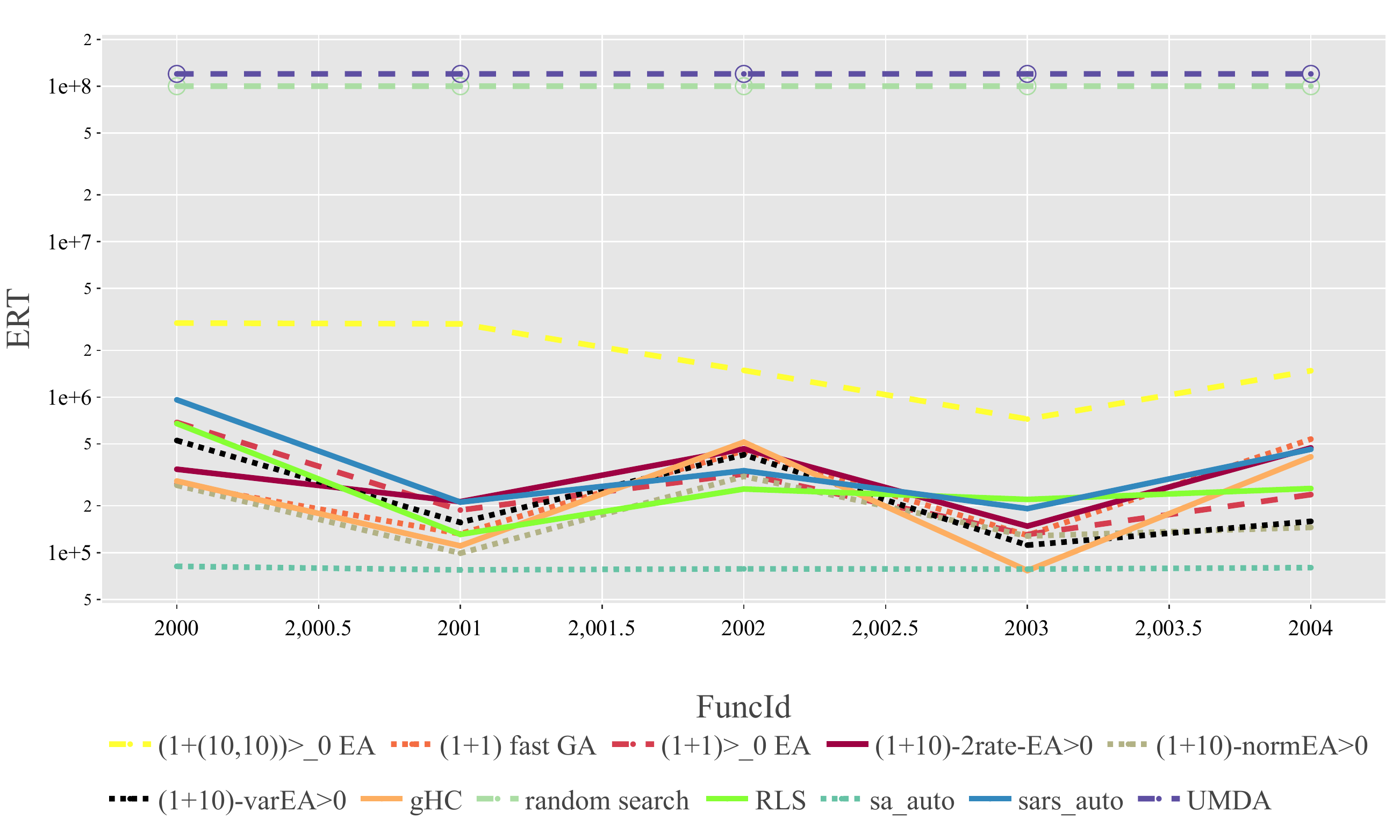}
    \caption{ERT values of 12 algorithms for the  $800$-dimensional maximum cut problem instances. The targets are selected to be the $0.02$ quantile of the final target (i.e., fitness) found by the best algorithm. Circles indicate that algorithms can not hit the corresponding target with the given budget.}
    \label{fig:ert_rank_maxcut}
\end{figure}

As shown in Figure~\ref{fig:ert_rank_maxcut}, random search and UMDA can not hit the corresponding target with the given budget though UMDA performs well for the maximum coverage functions. The $(1+(10,10))$~EA$_{>0}$ is also outperformed by the other EA variants and RLS significantly. The sa-auto presents the best result across all the problem instances, and apparently the restart strategy is not helpful for solving the maximum cut instances based on the results of the sars-auto. Note that the performance of gHC differs across the tested instances, for example, it ranks the best for the instance ``2003'' but is defeated by the other EA variants for the instance ``2002''. 

\begin{figure}
    \centering
    \includegraphics[width=0.48\textwidth]{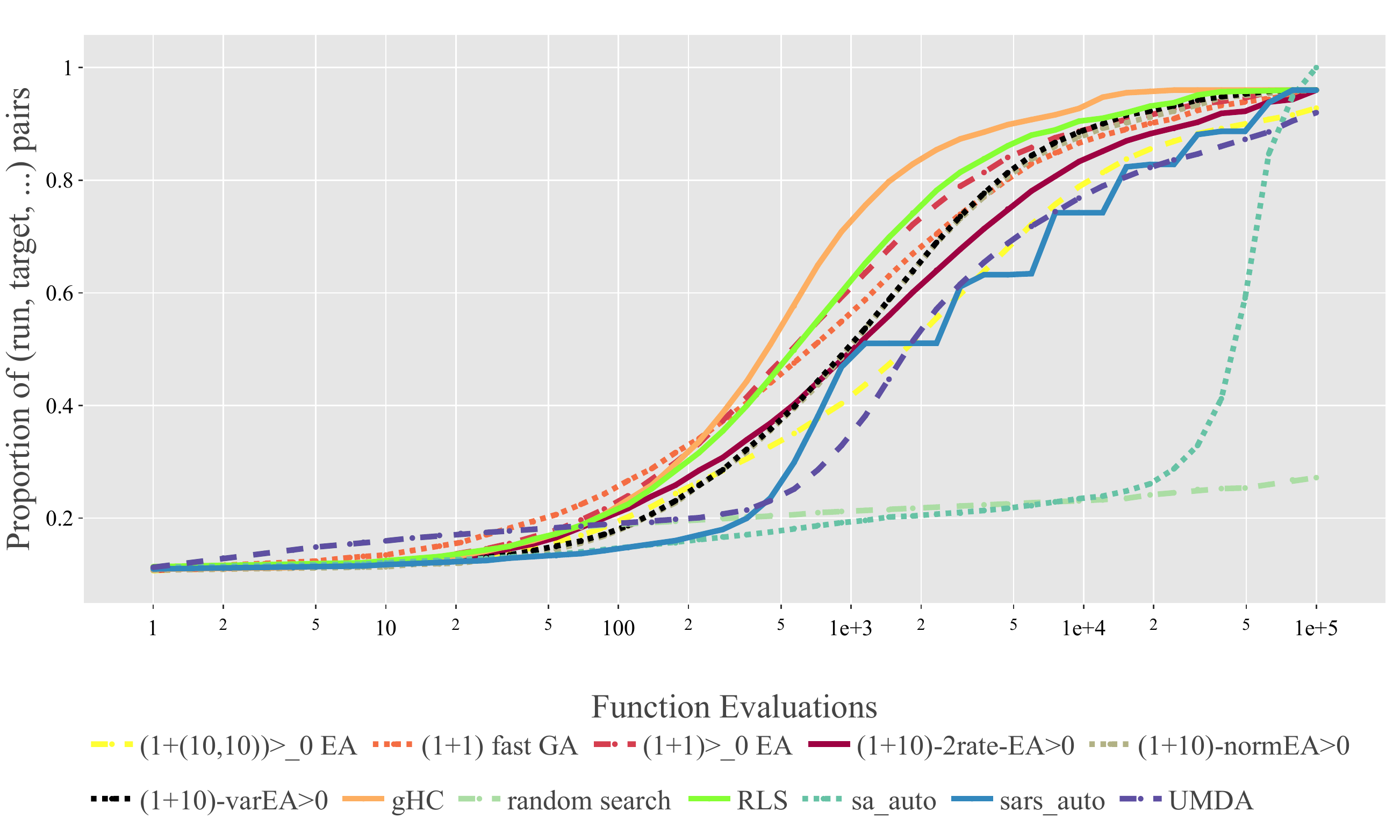}
    \caption{Empirical cumulative distribution function of 12 algorithms aggregated over all the $800$-dimensional maximum cut problem instances. Targets are 25 linearly spaced points between the best and worst found fitness values.}
    \label{fig:ecdf_maxcut}
\end{figure}

For the aggregated performance of multiple instances as shown in Figure~\ref{fig:ecdf_maxcut}, gHC coverges fast at the early stage of optimization again but it does not fall behind the other algorithms thereafter as shown in Figure~\ref{fig:ecdf}. In contrast, the sa-auto presents slow initial convergence, but it obtains better solutions than the other algorithms after using more function evaluations.

\begin{figure}
    \centering
    \includegraphics[width=0.48\textwidth]{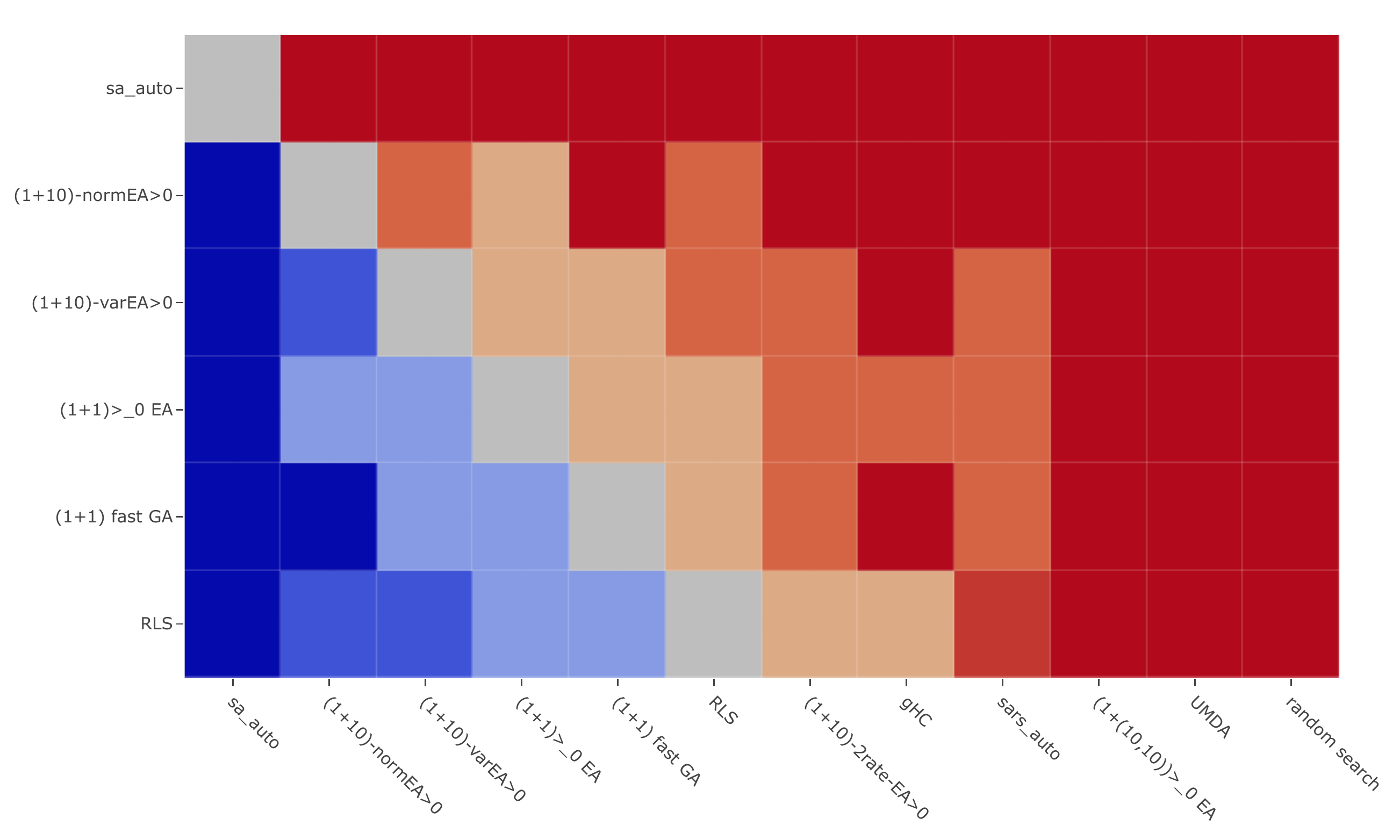}
    \caption{Heatmap of the fraction of times an algorithm's final obtained fitness value is better than the one achieved by another algorithm. The results are from comparisons over $5$ maximum cut problem instances.  }
    \label{fig:glico2_maxcut}
\end{figure}

Figure~\ref{fig:glico2_maxcut} plots a heatmap for the pairwise competitions of the algorithms across all the tested maximum cut problem instances. The color indicates the fraction of the times that one algorithm's (listed along $y$-axis) final fitness value is better than the one achieved by another algorithm (listed along $x$-axis). Blue indicates better results.
We list only the results of the comparisons with the $6$ best algorithms.
Following the ECDF results in Figure~\ref{fig:ecdf_maxcut}, we are not surprised to observe that the sa-auto wins the pairwise-comparison based ranking based on fixed-budget results. The EA variants and RLS show similar results and outperform UMDA and random search.


\section{Conclusions}

We have described a setup for benchmarking iterative search algorithms for submodular optimization problems. Different benchmark problems have been implemented and provided as part of IOHprofiler. The setup allows for a detailed comparison of different approaches to these benchmark problems and instances. We showcased this for the maximum coverage problem as a classical example of a monotone submodular problem and the well-known maximum cut problem as an example of a non-monotone and unconstrained submodular problem. The intention is to use the setup provided as part of upcoming competitions on submodular optimization at leading international conferences as well as in teaching activities as part of courses on heuristic search and evolutionary computation.

Based on our baseline of $12$ algorithms, the algorithms show  different performance across the two problem classes. For example, random search, gHC, RLS, and the $(1+10)$~varEA$_{>0}$ are outperformed by the other tested algorithms for the maximum coverage problem. However, the $(1+10)$~varEA$_{>0}$ and RLS show promising performance for the maximum cut problem. Moreover, we also observe performance variance across the instances of one problem. For example, the ranks of gHC and RLS regarding the ERT alter a lot for the instances of the maximum cut problem. Therefore, it would be interesting for future work to study the performance, e.g., convergence process, for particular problem instances, which can help us obtain insights into algorithms' behavior. Though we have tested a limited set of algorithms in this paper, the presented study cases provide us a good baseline for next steps such as investigating the impact of parameters for the evolutionary algorithms.

\section{Acknowledgements}
The authors thank Thomas Weise for the implementation of the simulated annealing algorithm. 
This work has been supported by the Australian Research Council (ARC) through grant FT200100536, and the Fundamental Research Funds for the Central Universities (0221-14380009)
\bibliographystyle{IEEEtran}
\bibliography{references} 

\begin{thebibliography}{10}
\providecommand{\url}[1]{#1}
\csname url@samestyle\endcsname
\providecommand{\newblock}{\relax}
\providecommand{\bibinfo}[2]{#2}
\providecommand{\BIBentrySTDinterwordspacing}{\spaceskip=0pt\relax}
\providecommand{\BIBentryALTinterwordstretchfactor}{4}
\providecommand{\BIBentryALTinterwordspacing}{\spaceskip=\fontdimen2\font plus
\BIBentryALTinterwordstretchfactor\fontdimen3\font minus
  \fontdimen4\font\relax}
\providecommand{\BIBforeignlanguage}[2]{{%
\expandafter\ifx\csname l@#1\endcsname\relax
\typeout{** WARNING: IEEEtran.bst: No hyphenation pattern has been}%
\typeout{** loaded for the language `#1'. Using the pattern for}%
\typeout{** the default language instead.}%
\else
\language=\csname l@#1\endcsname
\fi
#2}}
\providecommand{\BIBdecl}{\relax}
\BIBdecl

\bibitem{DBLP:books/cu/p/0001G14}
\BIBentryALTinterwordspacing
A.~Krause and D.~Golovin, ``Submodular function maximization,'' in
  \emph{Tractability: Practical approaches to hard problems}.\hskip 1em plus
  0.5em minus 0.4em\relax Cambridge University Press, 2014, pp. 71--104.
  [Online]. Available: \url{https://doi.org/10.1017/CBO9781139177801.004}
\BIBentrySTDinterwordspacing

\bibitem{Nemhauser:1978}
\BIBentryALTinterwordspacing
G.~L. Nemhauser and L.~A. Wolsey, ``Best algorithms for approximating the
  maximum of a submodular set function,'' \emph{Mathematics of Operations
  Research}, vol.~3, no.~3, pp. 177--188, 1978. [Online]. Available:
  \url{https://doi.org/10.1287/moor.3.3.177}
\BIBentrySTDinterwordspacing

\bibitem{vondrak2010submodularity}
J.~Vondr{\'a}k, ``Submodularity and curvature: The optimal algorithm,''
  \emph{{RIMS} K{\^{o}}ky{\^{u}}roku Bessatsu}, vol. B23, pp. 253–--266,
  2010.

\bibitem{DBLP:journals/ec/FriedrichN15}
\BIBentryALTinterwordspacing
T.~Friedrich and F.~Neumann, ``Maximizing submodular functions under matroid
  constraints by evolutionary algorithms,'' \emph{Evolutionary Computation},
  vol.~23, no.~4, pp. 543--558, 2015. [Online]. Available:
  \url{https://doi.org/10.1162/EVCO\_a\_00159}
\BIBentrySTDinterwordspacing

\bibitem{DBLP:conf/nips/QianYZ15}
\BIBentryALTinterwordspacing
C.~Qian, Y.~Yu, and Z.~Zhou, ``Subset selection by {P}areto optimization,'' in
  \emph{Proceedings of the 28th International Conference on Neural Information
  Processing Systems - Volume 1, NIPS 2015}, 2015, pp. 1774--1782. [Online].
  Available:
  \url{https://proceedings.neurips.cc/paper/2015/hash/b4d168b48157c623fbd095b4a565b5bb-Abstract.html}
\BIBentrySTDinterwordspacing

\bibitem{DBLP:conf/aaai/000100QR19}
\BIBentryALTinterwordspacing
T.~Friedrich, A.~G{\"{o}}bel, F.~Neumann, F.~Quinzan, and R.~Rothenberger,
  ``Greedy maximization of functions with bounded curvature under partition
  matroid constraints,'' in \emph{The 33rd {AAAI} Conference on Artificial
  Intelligence, {AAAI} 2019}.\hskip 1em plus 0.5em minus 0.4em\relax {AAAI}
  Press, 2019, pp. 2272--2279. [Online]. Available:
  \url{https://doi.org/10.1609/aaai.v33i01.33012272}
\BIBentrySTDinterwordspacing

\bibitem{DBLP:conf/aaai/Do021}
\BIBentryALTinterwordspacing
A.~V. Do and F.~Neumann, ``Pareto optimization for subset selection with
  dynamic partition matroid constraints,'' in \emph{{AAAI}}.\hskip 1em plus
  0.5em minus 0.4em\relax {AAAI} Press, 2021, pp. 12\,284--12\,292. [Online].
  Available: \url{https://ojs.aaai.org/index.php/AAAI/article/view/17458}
\BIBentrySTDinterwordspacing

\bibitem{DBLP:conf/ppsn/DoN20}
\BIBentryALTinterwordspacing
A.~Do and F.~Neumann, ``Maximizing submodular or monotone functions under
  partition matroid constraints by multi-objective evolutionary algorithms,''
  in \emph{{PPSN} {(2)}}, ser. Lecture Notes in Computer Science, vol.
  12270.\hskip 1em plus 0.5em minus 0.4em\relax Springer, 2020, pp. 588--603.
  [Online]. Available: \url{https://doi.org/10.1007/978-3-030-58115-2\_41}
\BIBentrySTDinterwordspacing

\bibitem{DBLP:journals/corr/abs-1911-11451}
\BIBentryALTinterwordspacing
B.~Doerr, C.~Doerr, A.~Neumann, F.~Neumann, and A.~M. Sutton, ``Optimization of
  chance-constrained submodular functions,'' in \emph{The Thirty-Fourth {AAAI}
  Conference on Artificial Intelligence, {AAAI} 2020}.\hskip 1em plus 0.5em
  minus 0.4em\relax {AAAI} Press, 2020, pp. 1460--1467. [Online]. Available:
  \url{https://www.aaai.org/Papers/AAAI/2020GB/AAAI-DoerrB.6164.pdf}
\BIBentrySTDinterwordspacing

\bibitem{DBLP:conf/ppsn/NeumannN20}
\BIBentryALTinterwordspacing
A.~Neumann and F.~Neumann, ``Optimising monotone chance-constrained submodular
  functions using evolutionary multi-objective algorithms,'' in \emph{Parallel
  Problem Solving from Nature, {PPSN} 2020, Proceedings, Part {I}}, ser. LNCS,
  vol. 12269.\hskip 1em plus 0.5em minus 0.4em\relax Springer, 2020, pp.
  404--417. [Online]. Available:
  \url{https://doi.org/10.1007/978-3-030-58112-1\_28}
\BIBentrySTDinterwordspacing

\bibitem{DBLP:journals/tec/LaumannsTZ04}
\BIBentryALTinterwordspacing
M.~Laumanns, L.~Thiele, and E.~Zitzler, ``Running time analysis of
  multiobjective evolutionary algorithms on pseudo-boolean functions,''
  \emph{{IEEE} Trans. Evol. Comput.}, vol.~8, no.~2, pp. 170--182, 2004.
  [Online]. Available: \url{https://doi.org/10.1109/TEVC.2004.823470}
\BIBentrySTDinterwordspacing

\bibitem{Giel2003}
O.~Giel, ``Expected runtimes of a simple multi-objective evolutionary
  algorithm.'' in \emph{Proceedings of the Congress on Evolutionary
  Computation, CEC 2003}, ser. IEEE Press, vol.~3, 2003, pp. 1918--1925.

\bibitem{DBLP:journals/corr/abs-2111-04077}
\BIBentryALTinterwordspacing
J.~de~Nobel, F.~Ye, D.~Vermetten, H.~Wang, C.~Doerr, and T.~B{\"{a}}ck,
  ``Iohexperimenter: Benchmarking platform for iterative optimization
  heuristics,'' \emph{CoRR}, vol. abs/2111.04077, 2021. [Online]. Available:
  \url{https://arxiv.org/abs/2111.04077}
\BIBentrySTDinterwordspacing

\bibitem{DBLP:journals/telo/WangVYDB22}
\BIBentryALTinterwordspacing
H.~Wang, D.~Vermetten, F.~Ye, C.~Doerr, and T.~B{\"{a}}ck, ``Iohanalyzer:
  Detailed performance analyses for iterative optimization heuristics,''
  \emph{{ACM} Trans. Evol. Learn. Optim.}, vol.~2, no.~1, pp. 3:1--3:29, 2022.
  [Online]. Available: \url{https://doi.org/10.1145/3510426}
\BIBentrySTDinterwordspacing

\bibitem{DBLP:conf/gecco/XieHAN019}
\BIBentryALTinterwordspacing
Y.~Xie, O.~Harper, H.~Assimi, A.~Neumann, and F.~Neumann, ``Evolutionary
  algorithms for the chance-constrained knapsack problem,'' in
  \emph{Proceedings of the Genetic and Evolutionary Computation Conference,
  {GECCO} 2019}.\hskip 1em plus 0.5em minus 0.4em\relax {ACM}, 2019, pp.
  338--346. [Online]. Available: \url{https://doi.org/10.1145/3321707.3321869}
\BIBentrySTDinterwordspacing

\bibitem{DBLP:conf/ppsn/NeumannXN22}
\BIBentryALTinterwordspacing
A.~Neumann, Y.~Xie, and F.~Neumann, ``Evolutionary algorithms for limiting the
  effect of uncertainty for the knapsack problem with stochastic profits,'' in
  \emph{{PPSN} {XVII} - 17th International Conference, {PPSN} 2022,
  Proceedings, Part {I}}, ser. Lecture Notes in Computer Science, vol.
  13398.\hskip 1em plus 0.5em minus 0.4em\relax Springer, 2022, pp. 294--307.
  [Online]. Available: \url{https://doi.org/10.1007/978-3-031-14714-2\_21}
\BIBentrySTDinterwordspacing

\bibitem{khuller1999budgeted}
S.~Khuller, A.~Moss, and J.~S. Naor, ``The budgeted maximum coverage problem,''
  \emph{Information processing letters}, vol.~70, no.~1, pp. 39--45, 1999.

\bibitem{DBLP:conf/kdd/KempeKT03}
\BIBentryALTinterwordspacing
D.~Kempe, J.~M. Kleinberg, and {\'{E}}.~Tardos, ``Maximizing the spread of
  influence through a social network,'' in \emph{Proceedings of the Ninth {ACM}
  {SIGKDD} International Conference on Knowledge Discovery and Data
  Mining}.\hskip 1em plus 0.5em minus 0.4em\relax {ACM}, 2003, pp. 137--146.
  [Online]. Available: \url{https://doi.org/10.1145/956750.956769}
\BIBentrySTDinterwordspacing

\bibitem{DBLP:reference/algo/Newman08}
A.~Newman, ``Max cut,'' in \emph{Encyclopedia of Algorithms}.\hskip 1em plus
  0.5em minus 0.4em\relax Springer, 2008.

\bibitem{DBLP:journals/eor/PolyakovskiyN17}
\BIBentryALTinterwordspacing
S.~Polyakovskiy and F.~Neumann, ``The packing while traveling problem,''
  \emph{Eur. J. Oper. Res.}, vol. 258, no.~2, pp. 424--439, 2017. [Online].
  Available: \url{https://doi.org/10.1016/j.ejor.2016.09.035}
\BIBentrySTDinterwordspacing

\bibitem{DBLP:conf/cec/BonyadiMB13}
\BIBentryALTinterwordspacing
M.~R. Bonyadi, Z.~Michalewicz, and L.~Barone, ``The travelling thief problem:
  The first step in the transition from theoretical problems to realistic
  problems,'' in \emph{Proceedings of the {IEEE} Congress on Evolutionary
  Computation, {CEC} 2013}, 2013, pp. 1037--1044. [Online]. Available:
  \url{https://doi.org/10.1109/CEC.2013.6557681}
\BIBentrySTDinterwordspacing

\bibitem{DBLP:conf/aaai/DoerrD0NS20}
\BIBentryALTinterwordspacing
B.~Doerr, C.~Doerr, A.~Neumann, F.~Neumann, and A.~M. Sutton, ``Optimization of
  chance-constrained submodular functions,'' in \emph{{AAAI}}.\hskip 1em plus
  0.5em minus 0.4em\relax {AAAI} Press, 2020, pp. 1460--1467. [Online].
  Available: \url{https://ojs.aaai.org/index.php/AAAI/article/view/5504}
\BIBentrySTDinterwordspacing

\bibitem{DBLP:books/cu/MotwaniR95}
R.~Motwani and P.~Raghavan, \emph{Randomized Algorithms}.\hskip 1em plus 0.5em
  minus 0.4em\relax Cambridge University Press, 1995.

\bibitem{DBLP:conf/wsdm/GoyalBL10}
\BIBentryALTinterwordspacing
A.~Goyal, F.~Bonchi, and L.~V.~S. Lakshmanan, ``Learning influence
  probabilities in social networks,'' in \emph{Proceedings of the Third
  International Conference on Web Search and Web Data Mining, {WSDM}
  2010}.\hskip 1em plus 0.5em minus 0.4em\relax {ACM}, 2010, pp. 241--250.
  [Online]. Available: \url{https://doi.org/10.1145/1718487.1718518}
\BIBentrySTDinterwordspacing

\bibitem{DBLP:journals/datamine/WangCW12}
\BIBentryALTinterwordspacing
C.~Wang, W.~Chen, and Y.~Wang, ``Scalable influence maximization for
  independent cascade model in large-scale social networks,'' \emph{Data Min.
  Knowl. Discov.}, vol.~25, no.~3, pp. 545--576, 2012. [Online]. Available:
  \url{https://doi.org/10.1007/s10618-012-0262-1}
\BIBentrySTDinterwordspacing

\bibitem{DBLP:journals/corr/abs-1812-00493}
\BIBentryALTinterwordspacing
E.~C. Pinto and C.~Doerr, ``Towards a more practice-aware runtime analysis of
  evolutionary algorithms,'' \emph{CoRR}, vol. abs/1812.00493, 2018. [Online].
  Available: \url{http://arxiv.org/abs/1812.00493}
\BIBentrySTDinterwordspacing

\bibitem{DBLP:conf/gecco/DoerrLMN17}
\BIBentryALTinterwordspacing
B.~Doerr, H.~P. Le, R.~Makhmara, and T.~D. Nguyen, ``Fast genetic algorithms,''
  in \emph{Proceedings of the Genetic and Evolutionary Computation Conference,
  {GECCO} 2017}.\hskip 1em plus 0.5em minus 0.4em\relax {ACM}, 2017, pp.
  777--784. [Online]. Available: \url{https://doi.org/10.1145/3071178.3071301}
\BIBentrySTDinterwordspacing

\bibitem{DBLP:journals/tcs/DoerrDE15}
\BIBentryALTinterwordspacing
B.~Doerr, C.~Doerr, and F.~Ebel, ``From black-box complexity to designing new
  genetic algorithms,'' \emph{Theor. Comput. Sci.}, vol. 567, pp. 87--104,
  2015. [Online]. Available: \url{https://doi.org/10.1016/j.tcs.2014.11.028}
\BIBentrySTDinterwordspacing

\bibitem{DBLP:conf/gecco/DoerrGWY17}
\BIBentryALTinterwordspacing
B.~Doerr, C.~Gie{\ss}en, C.~Witt, and J.~Yang, ``The (1+\emph{{\(\lambda\)}})
  evolutionary algorithm with self-adjusting mutation rate,'' in
  \emph{Proceedings of the Genetic and Evolutionary Computation Conference,
  {GECCO} 2017}.\hskip 1em plus 0.5em minus 0.4em\relax {ACM}, 2017, pp.
  1351--1358. [Online]. Available:
  \url{https://doi.org/10.1145/3071178.3071279}
\BIBentrySTDinterwordspacing

\bibitem{DBLP:conf/cec/YeDB19}
\BIBentryALTinterwordspacing
F.~Ye, C.~Doerr, and T.~B{\"{a}}ck, ``Interpolating local and global search by
  controlling the variance of standard bit mutation,'' in \emph{{IEEE} Congress
  on Evolutionary Computation, {CEC} 2019, Wellington, New Zealand, June 10-13,
  2019}.\hskip 1em plus 0.5em minus 0.4em\relax {IEEE}, 2019, pp. 2292--2299.
  [Online]. Available: \url{https://doi.org/10.1109/CEC.2019.8790107}
\BIBentrySTDinterwordspacing

\bibitem{nourani1998comparison}
Y.~Nourani and B.~Andresen, ``A comparison of simulated annealing cooling
  strategies,'' \emph{Journal of Physics A: Mathematical and General}, vol.~31,
  no.~41, p. 8373, 1998.

\bibitem{DBLP:journals/ec/Muhlenbein97}
\BIBentryALTinterwordspacing
H.~M{\"{u}}hlenbein, ``The equation for response to selection and its use for
  prediction,'' \emph{Evolutionary Computation}, vol.~5, no.~3, pp. 303--346,
  1997. [Online]. Available: \url{https://doi.org/10.1162/evco.1997.5.3.303}
\BIBentrySTDinterwordspacing

\bibitem{vevcek2014chess}
\BIBentryALTinterwordspacing
N.~Ve{\v{c}}ek, M.~Mernik, and M.~{\v{C}}repin{\v{s}}ek, ``A chess rating
  system for evolutionary algorithms: A new method for the comparison and
  ranking of evolutionary algorithms,'' \emph{Information Sciences}, vol. 277,
  pp. 656--679, 2014. [Online]. Available:
  \url{https://doi.org/10.1016/j.ins.2014.02.154}
\BIBentrySTDinterwordspacing

\end{thebibliography}

\end{document}